\documentclass[runningheads]{llncs}
\usepackage[T1]{fontenc}
\usepackage{graphicx}
\usepackage{amsfonts}
\usepackage{amsmath}
\usepackage{subcaption}
\usepackage{etoolbox}
\usepackage{kantlipsum}

\newcommand{\repthanks}[1]{\textsuperscript{\ref{#1}}}
\makeatletter
\patchcmd{\maketitle}
  {\def\thanks}
  {\let\repthanks\repthanksunskip\def\thanks}
  {}{}
\patchcmd{\@maketitle}
  {\def\thanks}
  {\let\repthanks\@gobble\def\thanks}
  {}{}
\newcommand\repthanksunskip[1]{\unskip{}}
\makeatother

\begin{document}
\title{PS-StyleGAN: Illustrative Portrait Sketching using Attention-Based Style Adaptation}
\titlerunning{PS-StyleGAN}
%
\author{Kushal Kumar Jain \thanks{equal contribution\protect\label{X}} \and
Ankith Varun J \repthanks{X} \and
Anoop Namboodiri}
%
%
\institute{IIIT-Hyderabad, Gachibowli, India  \\
\email{\{kushal.kumar, ankith.varun\}@research.iiit.ac.in and anoop@iiit.ac.in}\\}
\maketitle              

\begin{center}
    \centering
    \captionsetup{type=figure}\addtocounter{figure}{-1}
    \begin{subfigure}{0.159\textwidth}
        \includegraphics[width=\linewidth]{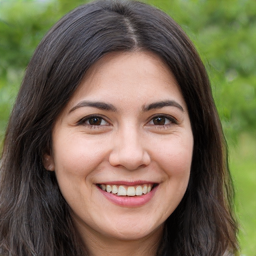}
        \caption{Identity}
        \label{fig:identity}
    \end{subfigure}
    \begin{subfigure}{0.159\textwidth}
        \includegraphics[width=\linewidth]{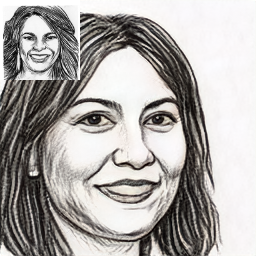}
        \caption{Smile3}
        \label{fig:fs2k_1}
    \end{subfigure}
    \begin{subfigure}{0.159\textwidth}
        \includegraphics[width=\linewidth]{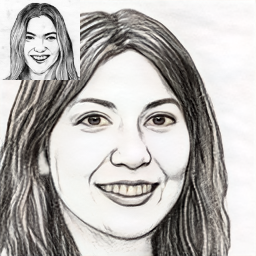}
        \caption{Smile2}
        \label{fig:fs2k_2}
    \end{subfigure}
    \begin{subfigure}{0.159\textwidth}
        \includegraphics[width=\linewidth]{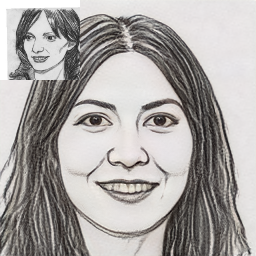}
        \caption{Smile2}
        \label{fig:fs2k_3}
    \end{subfigure}
    \begin{subfigure}{0.159\textwidth}
        \includegraphics[width=\linewidth]{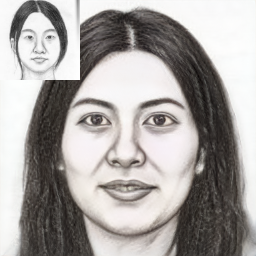}
        \caption{Smile1}
        \label{fig:cuhk}
    \end{subfigure}
    \begin{subfigure}{0.159\textwidth}
        \includegraphics[width=\linewidth]{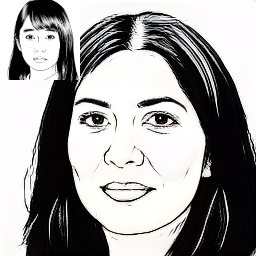}
        \caption{Neutral}
        \label{fig:apd}
    \end{subfigure}
    \captionof{figure}{Outputs of PS-StyleGAN for different sketching styles (inset) in specified poses and expressions while maintaining the input identity. A model trained on FS2K dataset was used for (b) - (d), while CUHK and APDrawing were used for the models in (e) and (f).}
    \label{illus}
\end{center}

\begin{abstract}
Portrait sketching involves capturing identity specific attributes of a real face with abstract lines and shades. Unlike photo-realistic images, a good portrait sketch generation method needs selective attention to detail, making the problem challenging. This paper introduces \textbf{Portrait Sketching StyleGAN (PS-StyleGAN)}, a style transfer approach tailored for portrait sketch synthesis. We leverage the semantic $W+$ latent space of StyleGAN to generate portrait sketches, allowing us to make meaningful edits, like pose and expression alterations, without compromising identity. To achieve this, we propose the use of Attentive Affine transform blocks in our architecture, and a training strategy that allows us to change StyleGAN's output without finetuning it. These blocks learn to modify style latent code by paying attention to both content and style latent features, allowing us to adapt the outputs of StyleGAN in an inversion-consistent manner. Our approach uses only a few paired examples ($\sim 100$) to model a style and has a short training time. We demonstrate PS-StyleGAN's superiority over the current state-of-the-art methods on various datasets, qualitatively and quantitatively.

\end{abstract}
\section{Introduction}
\label{sec:intro}
\begin{figure}[t!]
    \centering
    \begin{subfigure}{0.24\columnwidth}
        \includegraphics[width=\linewidth]{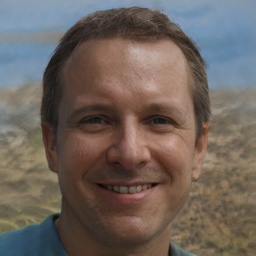}
        \caption{input}
        \label{fig:content}
    \end{subfigure}
    \begin{subfigure}{0.24\columnwidth}
        \includegraphics[width=\linewidth]{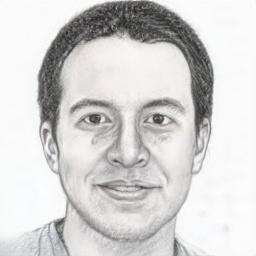}
        \caption{sketch}
        \label{fig:sketch}
    \end{subfigure}
    \begin{subfigure}{0.24\columnwidth}
        \includegraphics[width=\linewidth]{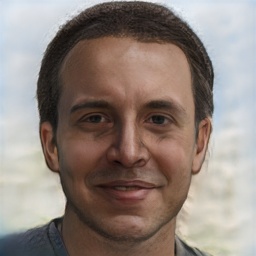}
        \caption{structure only}
        \label{fig:structure}
    \end{subfigure}
    \begin{subfigure}{0.24\columnwidth}
        \includegraphics[width=\linewidth]{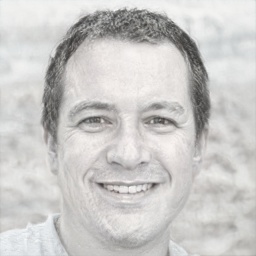}
        \caption{color only}
        \label{fig:color}
    \end{subfigure}
    \caption{Results of DualStyleGAN trained on CUHK \cite{cuhk} dataset. The generated sketch (b) is a result of complete structure and color transfer. Structure transfer (c) results in considerable loss of identity while color transfer (d) does not yield stylization.}  
    \label{fig:dualstylegan}
\end{figure}

Drawing portrait sketches is an intricate but timeless form of artistic expression. It requires one to use minimalistic elements, such as lines, to encapsulate the distinctive features and the overall essence of an individual's identity. A lot of research has been done to understand how humans perceive 3D shapes through rough sketches or simple line drawings and why they effectively represent complex concepts like identity \cite{SurfaceVE,HertzmannWhyDL}. While some theories exists, it is still not very well known as to how artists choose the lines that they draw \cite{HertzmannROLE,Linesasedges}. Hence the process of generation of such sketches remain a manual creation and time-consuming one. Many Non-Photorealistic Rendering methods tried to solve this artistic challenge \cite{DeCarlo2003SuggestiveCF,NPRLineDF,OhtakeRidge,Hertsmann_stroke_3d,style_abstraction}. However, they rely on ground truth geometry, which is noisy near detailed parts of the face like eyes, nose and lips. Humans are particularly sensitive to details in these regions as we have dedicated neural pathways \cite{MechanismsOF} for face detection and identification. \\
Deep learning \cite{VeryDC} based approaches like style transfer \cite{Gatys2016ImageST} and image to image translation \cite{pix2pix2017,cyclegan} have been very successful in sketch generation. 
Innovations in style transfer \cite{Ulyanov2016texture,AdaIN,SANet} have made the generation process faster and more reliable. 
However, these methods only perform well for global texture transformations and do not consider local details, abstracting out crucial elements like eyes and lips. 
Following the development of Generative Adversarial Networks (GANs) \cite{Goodfellow2014GenerativeAN}  and cGANs \cite{cGAN}, Isola {\it et al.} proposed a novel method for general image to image translation using cGANs \cite{pix2pix2017,cyclegan}. 
Even though training such generators is notoriously difficult, modern image to image translation methods \cite{pix2pixsketch1} have shown impressive results and tremendous potential.\\
More recently, researchers have used a pretrained StyleGAN \cite{Karras_2019_CVPR,Karras2018ASG} along with encoders that invert a given image into StyleGAN's latent space to tackle the problem of general image to image translation \cite{EditingIS}. 
Its highly semantic $W+$ latent space allows one to make meaningful edits to the final output, like changing pose, facial expression or emotion, without affecting the identity. 
However, for portrait sketch generation, incorporating the sketch style into the original StyleGAN poses a significant challenge. It carries the risk of perturbing and changing the behaviour of the latent space of the pre-trained StyleGAN, making latent editing difficult. Yet a lot of works \cite{StyleTF,TransEditor,TransformingTL,DualStyleGAN,JoJoGan}, have tried to solve the issue with different approaches. Some methods \cite{DualStyleGAN,TransEditor} divide the latent space into dual spaces and others use attention \cite{TransformingTL} for better mapping features in the latent space. \\
DualStyleGAN \cite{DualStyleGAN} achieves portrait style transfer by disentangling the spatial resolution layers of StyleGAN to perform independent structure and color transfer between domains. They propose a ResBlock \cite{ResNet} based feature statistics alignment module using AdaIN \cite{AdaIN} that incorporates \textit{structure control} over the coarse and middle layers of StyleGAN. Training their ResBlock does not change the latent distribution and thus it allows semantic editing. However, DualStyleGAN's style blending might result in significant loss of identity. In our experiments we observed some extent of structure and color entanglement across all layers of StyleGAN, especially in the middle layers. Hence, it is difficult to decouple structure and color transformations without losing desirable artistic characteristics like pencil strokes and shading, see Fig \ref{fig:dualstylegan}. Lastly, DualStyleGAN relies on a pre-training process to learn structure transfer in the source domain which is quite time consuming. \\
To this end, we propose Portrait Sketching StyleGAN (PS-StyleGAN), which converts real face photos into a portrait sketch while offering the semantic editability of StyleGAN without the need to finetune it. We use our novel attention \cite{Attention} based affine transformation blocks to modulate only the style latent codes while keeping the generator frozen. These blocks help us simulate the behaviour of a finetuned StyleGAN. We discard any form of structure transfer so as to ensure identity preservation and adopt a progressive training strategy to achieve a rapid but smooth domain transfer. We also run our model on different datasets to show that our model is inversion consistent. Our main contributions are: \begin{enumerate}
    \item We propose PS-StyleGAN, which can generate expressive portrait sketches from a photo-realistic face image. Specifically, our method can learn complex hairstyles and generate perfect eyes, nose and lips while preserving the subtleties of an artist's style. Furthermore, our model converges quickly and can be trained on relatively small datasets.
    \item We introduce a novel Attentive Affine transformation for better-transforming style latent codes based on style examples. 
    \item We perform experiments, conduct a user study and run ablations on various datasets to show the effectiveness of our method.
\end{enumerate}

\section{Related Works}
\label{sec:related}
\subsection{Image To Image Translation}
Image-to-image translation techniques aim to learn a mapping function that can convert an input image from one domain into the corresponding image in another domain. This approach was initially introduced by Isola {\it et al.} using conditional GANs \cite{pix2pix2017} and has since seen significant development. Recent methods like \cite{HIDA} use Dynamic Normalization (DySPADE) in the generator architecture along with depth maps to supervise the generation with encouraging results. An unsupervised version of Pix2Pix called the CycleGAN \cite{cyclegan,UNIT,UGATIT} sparked the creation of some fascinating sketch generation methods. In AP-DrawingGAN, Yi {\it et al.} \cite{APDrawingGAN2019} used dedicated GANs to generate difficult-to-sketch features like eyes, nose and lips. FSGAN \cite{FS2K} extends their approach and introduces a new dataset called FS2K, which has three styles and paired sketch examples. We use this dataset for training and comparison with other methods. New approaches like \cite{pix2pixsketch1,jain2024clip4sketchenhancingsketchmugshot}, use CLIP \cite{clip} embeddings. In \cite{pix2pixsketch1} the authors use CLIP along with a geometry-preserving loss to achieve line drawings that respect the scene's geometry. These methods require training the generator, which is difficult and necessitates large datasets, which is not feasible for face sketches.\\
Diffusion models \cite{Ho2020DenoisingDP,dhariwal2021diffusion} have made significant progress in text-guided image generation \cite{Ramesh2022HierarchicalTI,Saharia2022PhotorealisticTD,Rombach2021HighResolutionIS}, in the past few years. Personalised sketch generation in the context of diffusion models has been achieved by either finetuning the generator itself \cite{Ruiz2022DreamBoothFT} or by learning personalised word or image embeddings for the generator \cite{Gal2022TI,Ye2023IPAdapterTC}, or by using ID embeddings as condition \cite{li2023photomaker,jain2024clip4sketchenhancingsketchmugshot}.  
Our method instead relies on a StyleGAN generator trained on realistic human faces \cite{Karras2018ASG}. We modify the generator's output by learning crucial aspects of each style using only a few examples.
\vspace*{-\baselineskip}
\subsection{StyleGAN Latent Space Inversion}
The exceptional image quality and semantic richness of StyleGAN \cite{Karras2018ASG,Karras_2019_CVPR} has made it very attractive for directed image generation. GANs synthesize images by sampling a vector (latent code) from the latent space distribution. GAN inversion tackles the problem of finding the latent code that best recreates a given image. This can be done by direct optimization, learning encoders or a mix of both \cite{Abdal_2019_ICCV,EditingIS}. For latent editing, some methods take the supervised approach by finding latent directions for labelled attributes, while others take a more unsupervised approach \cite{Unsupervised_editing,E4E}. \\
Although there are many approaches for latent space manipulation of realistic images, they do not work for stylized generators 
as it would change the latent distribution, making latent manipulation inconsistent \cite{AgileGAN}. JoJoGAN \cite{JoJoGan} achieves style transfer by training a new mapper for every finetuned StyleGAN, but it comes at the cost of identity.
DualStyleGAN \cite{DualStyleGAN} solves this issue by disentangling the spatial resolution layers of StyleGAN to perform independent structure and color transfer between domains. They propose a ResBlock \cite{ResNet} based feature statistics alignment module using AdaIN \cite{AdaIN} that incorporates \textit{structure control} over the coarse and middle layers of StyleGAN. Training their ResBlock does not change the latent distribution and thus it allows semantic editing. Our method differs from their approach as we investigate how to better preserve content structure, as portrait sketches have well-defined lines that need spatial consistency. We use attention based style adaption blocks to smoothly transform the generative space by aligning it to the feature statistics of the style examples.
\vspace*{-\baselineskip}
\subsection{Attention in Latent Space Manipulation}
Following the development of $W$+ space \cite{Abdal_2019_ICCV} many methods have tried to find new latent spaces that can offer better reconstruction ability while retaining the editing ability. In StyleTransformer \cite{StyleTF}, the authors use cross and self attention layers to aid in the inversion task of $W$+ latent space, showing that transformers can be a useful addition here. In  TransStyleGAN \cite{TransformingTL}, the authors introduce a new $W$++ latent space by replacing the MLP layers in the mapper network with transformer layers, resulting in better reconstruction and editing abilities.DualStyleGAN \cite{DualStyleGAN} proposes splitting the latent space into dual spaces effectively disentangling style and content spaces, allowing existing editing approaches to work on stylized spaces. A similar approach dubbed TransEditor \cite{TransEditor} also divides the latent space into $P$ and $Z$ spaces but crucially also uses cross attention based interaction module to correlate between the separated spaces. In \cite{DisentangledIG}, the authors find that editing the style codes in early stages of the generation process affects the structural properties of the image, resulting in artefacts in the final results. TransEditor mitigates this issue by increasing collaboration between the two spaces. In our approach we use attention based style adaption blocks to transform the style codes only in the later stages of the generation process.

\section{Method}
\label{sec:method}
\begin{figure}[t!]
    \centering
    \includegraphics[width=\columnwidth]{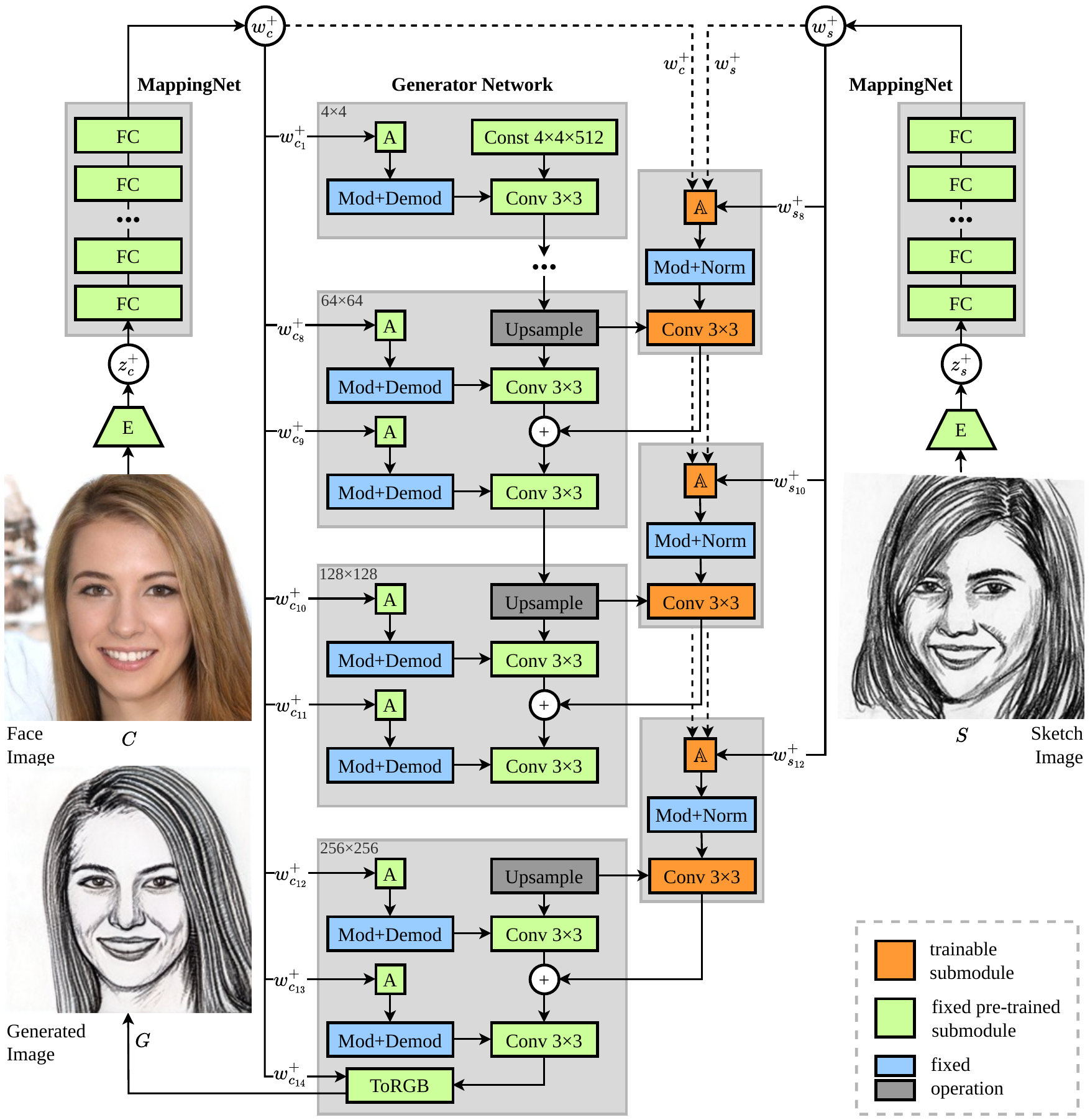}
    \caption{An overview of our model architecture. We use a pretrained 256x256 resolution StyleGAN2 \cite{Karras_2019_CVPR} generator $g$ fitted with three style adaptation blocks at the fine resolution layers. Each block consists of a novel Attentive Affine transform module ($\mathbb{A}$) that predicts affine parameters from attention-weighted latent codes of $S$ using supervision from $w_c^+$ and $w_s^+$. These parameters are then used to modulate and normalize the spatial features of $g$ at different scales to imbibe the style $\mathbb{S}$ into $C$.}
    \label{modelarch}
\end{figure}

We propose an end-to-end method for facial sketch synthesis using our model PS-StyleGAN $g'$, whose architecture is outlined in Fig. \ref{modelarch}. Given a content image $C$ and sketch image $S$ of a particular style $\mathbb{S}$, we invert both images to the $Z+$ latent space of a pre-trained StyleGAN generator $g$ using a pSp-based encoder $E$ \cite{Richardson2020EncodingIS,E4E}. We train the encoder $E$ on $256\times256$ resolution of FFHQ dataset \cite{Karras2018ASG} and modify it to embed face images to the $Z+$ latent space, which is more resilient to background details than the standard $W+$ space as observed in \cite{AgileGAN}. Using StyleGAN's mapping network $f$, we transform them into latent codes $w_c^+$ and $w_s^+$, respectively, in the shared $W+$ latent space of $g$. Finally, we pass the latent codes through our novel synthesis network $g'$ to obtain the generated image $G$, successfully capturing the style of $\mathbb{S}$. In the following sections, we give a detailed description of our model architecture and training procedure.

\subsection{Hierarchical Style Control in StyleGAN}
\label{sec:3.1}
 As described in \cite{Karras2018ASG}, the style blocks/layers of StyleGAN of different spatial resolutions controlled specific aspects of face generation. \textit{Coarse layers} ($4\times4 - 8\times8$ resolution) affect high-level aspects such as pose, hair texture, face structure and accessories. \textit{Middle layers} ($16\times16 - 32\times32$ resolution) generate smaller-scale features like eyes, smile, hairstyle, etc. \textit{Fine layers} ($64\times64 - 256\times256$ resolution) mainly control the general color scheme and microstructure of the generated image.

To tackle the challenges pointed out in Sec \ref{sec:intro}, we use attention-based style adaptation blocks in the fine layers of the generator network that perform feature transformations by considering both global and local style patterns. Each block consists of a novel Attentive Affine transform module ($\mathbb{A}$) and StyleGAN's modulative convolution layer, which provide instance-wise style conditioning to the content features. We choose to modulate just the fine layer features of the generator so as to preserve the overall structure of the content image. The adapted features are then fused with the original content features at each layer to allow a smooth transition of the generative space from the photo-realistic domain to the sketch domain. We show experimentally in Sec 3.1 of the supplementary that the latent space of StyleGAN remains consistent, allowing us to manipulate sketches using methods designed for realistic images.

We use the following notation for subsequent analysis - $w_{x_i}^+$ denotes the $i^\text{th}$ segment of the latent code of an input image $X$,  $F_i^X$ denotes the feature maps of $X$ that go into the $i^\text{th}$ convolution layer of the synthesis network and $y_i^X=(y_{s,i}^X,y_{b,i}^X)$ denotes the corresponding affine parameters computed at that layer.

\subsection{Paying Attention in Latent Space}
\label{sec:3.2}

\begin{figure*}[t]
    \centering
    \begin{subfigure}{0.275\textwidth}
        \includegraphics[width=\linewidth]{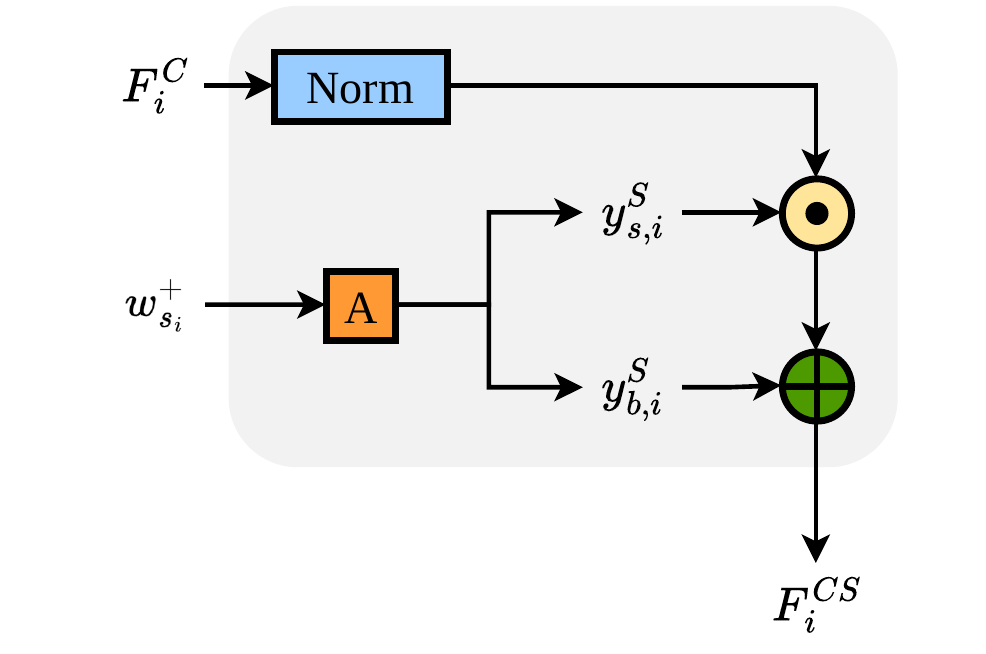}
        \caption{AdaIN}
        \label{fig:AdaIN}
    \end{subfigure}
    \begin{subfigure}{0.355\textwidth}
        \includegraphics[width=\linewidth]{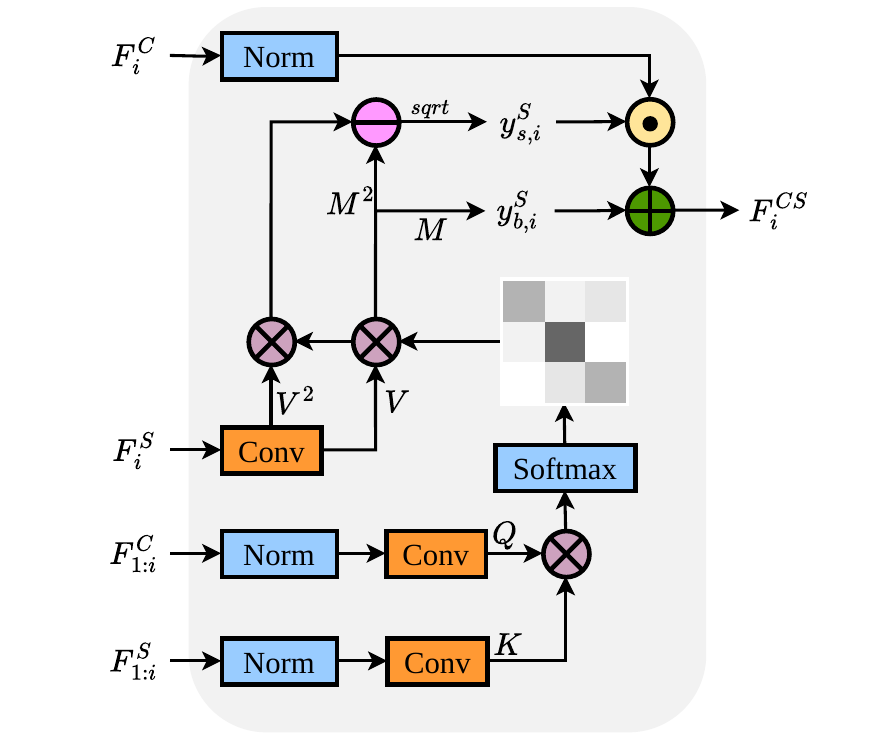}
        \caption{AdaAttN}
        \label{fig:AdaAttN}
    \end{subfigure}
    \begin{subfigure}{0.345\textwidth}
        \includegraphics[width=\linewidth]{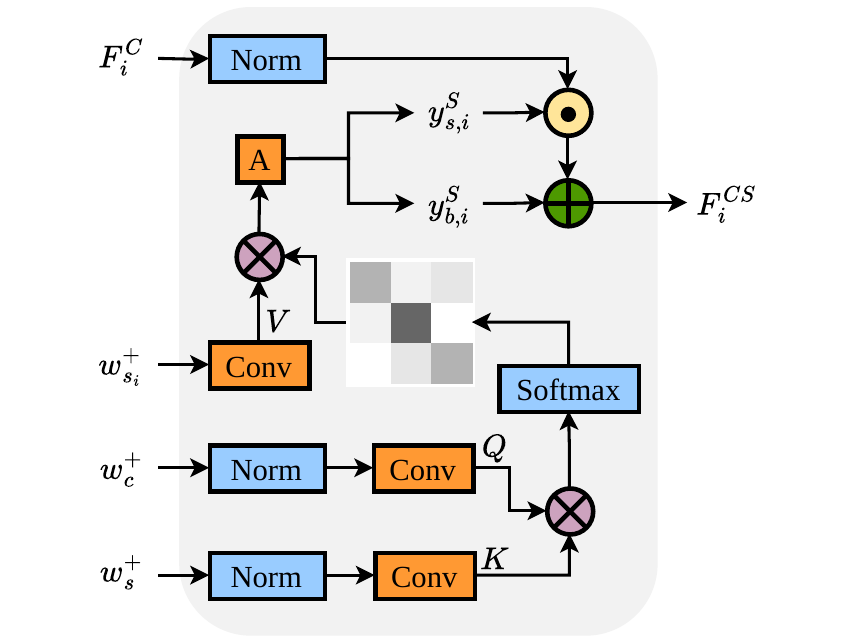}
        \caption{Ours}
        \label{fig:ours}
    \end{subfigure}
    \caption{(a) The structure of AdaIN \cite{AdaIN} module used in StyleGAN \cite{Karras2018ASG}. (b) The structure of AdaAttN \cite{AdaAttN} module. (c) The structure of our proposed design showing \textit{attentive affine} transform blocks. Here, \textit{A} denotes a basic affine transform block consisting of a single trainable fully-connected layer and \textit{Norm} denotes channel-wise mean-variance normalization.}  
    \label{fig:AffineArch}
\end{figure*}

Inspired by AdaAttN \cite{AdaAttN}, we introduce \textit{attentive affine} transformations to obtain improved affine parameters $y_i^S$ at the fine layers, which encapsulate the complete feature distribution of the style image. These parameters are then used by the AdaIN operation to achieve style transfer. As shown in Fig \ref{fig:ours}, the style adaptation process works in three steps.

\begin{enumerate}
    \item Computing attention maps with content and style latent codes $w_c^+$ and $w_s^+$, respectively.
    \item Calculating weighted segment of the style latent code and obtaining improved affine parameters $y_{s,i}^S$ and $y_{b,i}^S$ of the style features.
    \item Adaptively normalizing the content features for instance-wise feature distribution alignment.
\end{enumerate}

{\bfseries Attention Map Generation:} Different from standard style transfer methods, we use the attention mechanism to measure the similarity between the content and style latent codes instead of the corresponding features themselves. Due to the highly disentangled nature of the $W+$ latent space of StyleGAN, similarity in the latent space extrapolates well to that in the feature space. The relatively low dimensionality of the latent space keeps the model lightweight and cuts down on the computational costs of calculating attention maps. To compute the attention map $A$ corresponding to the fine layer $i$, we formulate query ($Q$), key ($K$) and value ($V$) as given below.
\begin{align*}
    Q&=f(\textit{Norm}(w_c^+)) \notag \\
    K&=g(\textit{Norm}(w_s^+)) \tag{1} \\
    V&=h(w_{s_i}^+) \notag
\end{align*}
where $f$, $g$, and $h$ are standard trainable $1 \times 1$ convolution layers while \textit{Norm} is the instance normalization operation carried out channel-wise. We compute attention map $A$ as:
\begin{equation}
    A=\textit{Softmax}(Q^T \otimes K) \tag{2}
\end{equation}
where $\otimes$ represents matrix multiplication.

{\bfseries Improved Affine Parameters:} In AdaAttN \cite{AdaAttN}, applying the attention map to the style feature $F_i^S$ is interpreted as observing a target style feature point as a distribution of all the weighted style feature points by attention. Then, statistical parameters are calculated from each distribution for subsequent modulation. In our case, the style latent code segment $w_{s_i}^+$ is multiplied with the attention score matrix to represent it as a distribution of all style points in the latent space. We term this as \textit{attention-weighted} latent code segment $x_{s_i}^+ \in \mathbb{R}^{512}$ from which we learn improved affine transformations to get better representative affine parameters $y_{s,i}^S$ and $y_{b,i}^S$ as follows.
\begin{equation}
     x_{s_i}^+=V \otimes A^T \tag{3}
\end{equation}
\begin{equation}
    (y_{s,i}^S,y_{b,i}^S)=\textit{Affine}(x_{s_i}^+) \tag{4}
\end{equation}
where \textit{Affine} is a learnable single fully-connected layer identical to the traditional StyleGAN's affine transform. The output dimensionality of the layer is twice the number of feature maps on the corresponding spatial resolution of the generator.

{\bfseries Adaptive Normalization:} Finally, we use the obtained affine parameters to modulate the normalized content feature map point-wise for each channel to generate the transformed feature map. Thus, the AdaIN operation in our case would become
\begin{equation}
F_i^{CS} = y_{s,i}^S \frac{F_i^C - \mu(F_i^C)}{\sigma(F_i^C)} + y_{b,i}^S \tag{5} \label{eq:5}
\end{equation}
The transformed feature maps $F_i^{CS}$ go into a trainable convolution layer whose outputs are selectively fused with those of the fine layers of the pre-trained synthesis network $g$ to complete the style adaptation process. We notice that omission of the mean affine parameter i.e $y_{b,i}^S$ during modulation does not affect the generated results. Therefore, like StyleGAN2 \cite{Karras_2019_CVPR}, we combine the modulation and convolution operation by scaling the convolution weights and effectively reduce the output dimensionality of the affine transform blocks.\\
To summarize, we perform feature statistics alignment using \textit{attentive affine} transformations by generating attention-weighted latent code that better represents the target style feature distribution in the fine layers ensuring that middle and coarse layer features are not lost.
\subsection{Training Strategy}
\begin{figure}[t!]
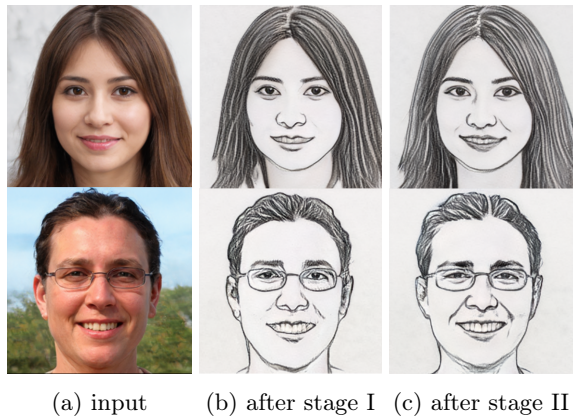

    \centering
    \begin{subfigure}{0.2\columnwidth}
        \includegraphics[width=\linewidth]{imgs/train_input.png}
        \caption{input}
        \label{fig:train_input}
    \end{subfigure}
    \begin{subfigure}{0.198\columnwidth}
        \includegraphics[width=\linewidth]{imgs/train_1000_iters.png}
        \caption{after stage I}
        \label{fig:train_1000_iters}
    \end{subfigure}
    \begin{subfigure}{0.2\columnwidth}
        \includegraphics[width=\linewidth]{imgs/train_2000_iters.png}
        \caption{after stage II}
        \label{fig:train_2000_iters}
    \end{subfigure}
    \caption{Results after each stage of progressive transfer learning. At the end of stage I, the model converges to an average representative style as seen in (b) where the eyes, nose and mouth are sketched in a similar manner. Stage II widens the model's generative space to capture subtle style variations resulting in better identity preservation as shown in (c).}  
    \label{fig:train}
\end{figure}

We adopt a progressive transfer learning scheme using a pretrained StyleGAN to smoothly refine its generative space to align with the target style distribution $\mathbb{S}$ comprising of limited samples. The scheme consists of two stages as illustrated in Fig \ref{fig:train}.\\
{\bfseries Stage I - Domain Transfer:} Similar to fine-tuning, we seek to achieve a general transformation from the photo-realistic domain to the sketch domain defined by $\mathbb{S}$. We randomly generate a latent code $z+$ and sample a sketch image $S$ and its corresponding style latent code $z_s^+$. Using StyleGAN's mapping network $f$, we obtain the $W+$ latent space embeddings for the content and style images as $w^+=f(z^+)$ and $w_s^+=f(z_s^+)$, respectively. Subsequently, we pass the latent codes through our synthesis network $g'$ to obtain the generated sketch as $G=g'(w^+,w_s^+)$. Following standard style transfer practices, we employ a style loss to fit the style of the generated sketch $G$ to $S$ which is given by
\begin{equation}
\mathcal{L}_{\text{sty}}=\lambda_{\text{CX}}\mathcal{L}_{\text{CX}}(G,S)+\lambda_{\text{FM}}\mathcal{L}_{\text{FM}} (G,S) \tag{6}
\end{equation}
where $\mathcal{L}_\text{CX}$ denotes contextual loss \cite{contextual} and $\mathcal{L}_\text{FM}$ denotes feature matching loss \cite{AdaIN}. To preserve the content features we use an identity loss \cite{ArcFaceAA} between $G$ and the reconstructed content image $g(w^+)$ thus constituting a content loss as follows.
\begin{equation}
\mathcal{L}_{\text{cont}}=\lambda_{\text{ID}}\mathcal{L}_{\text{ID}}(G,g(w^+)) \label{eq:7} \tag{7}
\end{equation}
where $\mathcal{L}_{\text{ID}}$ represents identity loss. Adding the standard StyleGAN adversarial loss $\mathcal{L}_{\text{adv}}$, our complete objective function takes the form of
\begin{equation}
\underset{G}{\text{min}} \hspace{2pt} \underset{D}{\text{max}} \hspace{2pt} \lambda_{\text{adv}} \mathcal{L}_{\text{adv}} + \mathcal{L}_{\text{sty}} + \mathcal{L}_{\text{cont}}  \notag
\end{equation}

{\bfseries Stage II - Conditional Refinement:} Stage I transforms StyleGAN's generative space to a narrow domain, failing to capture the diversity of styles contained in $\mathbb{S}$ as shown in Fig \ref{fig:train_1000_iters}. We use paired data of ground truth sketches and their photo-realistic counterparts as conditional supervision to broaden the generative domain. Given a sketch image $S$ and corresponding photo $P$, we get the $W+$ latent space embeddings as $w_s^+=f(E(S))$ and $w_p^+=f(E(P))$, and use them to obtain the generated sketch $G=g'(w_p^+,w_s^+)$. In addition to the losses used in stage I, we use perceptual loss \cite{perceptual} for $G$ to reconstruct $S$ thereby learning a varied set of style specific transformations. We also introduce a regularization term in $\mathcal{L}_{\text{cont}}$ which is the $L_2$ norm of the convolution weights comprising our style adaptation blocks. Therefore, Eq \ref{eq:7} changes to
\begin{equation}
\mathcal{L}_{\text{cont}}=\lambda_{\text{ID}}\mathcal{L}_{\text{ID}}(G,g(w_p^+)) + \lambda_{\text{reg}} ||W||_2 \tag{8}
\end{equation}
where $W$ represents the weight matrices of the trainable convolution layers. This regularization term controls the degree of style adaptation and helps prevent overfitting. Thus, the objective function modifies to
\begin{equation}
\underset{G}{\text{min}} \hspace{2pt} \underset{D}{\text{max}} \hspace{2pt} \lambda_{\text{adv}} \mathcal{L}_{\text{adv}} + \lambda_{\text{perc}} \mathcal{L}_{\text{perc}} + \mathcal{L}_{\text{sty}} + \mathcal{L}_{\text{cont}}   \notag
\end{equation}

\section{Experiments}

In this section, we assess the effectiveness of our proposed method by conducting comprehensive evaluations, which include qualitative and quantitative comparisons. \\
\textbf{Datasets :} We carry out our experiments on the FS2K dataset \cite{FS2K}, which stands as the most extensive publicly available FSS (Face Sketch Synthesis) dataset to date. This dataset comprises a substantial collection of 2,104 photo-sketch pairs, featuring a wide diversity of image backgrounds, skin tones, sketch styles, and lighting conditions. These sketches are classified into three distinct artistic styles. We also use the CUHK dataset \cite{cuhk}, which comprises mostly of asian faces, to measure our method against DualstyleGAN, a technique that introduces a bias of shape characteristics within the results. We further experiment with AP-Drawing dataset \cite{APDrawingGAN2019} to evaluate our method's ability to generalize and adapt to challenging sketching scenarios. \\
\textbf{Comparison methods:} We compare our method to other state of the art methods that have shown good performance in facial sketch synthesis, like HIDA \cite{HIDA}, FSGAN \cite{FS2K}, DualStyleGAN \cite{DualStyleGAN} and AdaAttN \cite{AdaAttN}. 

\subsection{Quantitative Analysis}

To quantitatively compare our method with others, we utilize four performance metrics: Learned Perceptual Image Patch Similarity (LPIPS) \cite{lpips}, Structure Co-Occurrence Texture (SCOOT) \cite{Scoot}, Feature Similarity Measure (FSIM) \cite{FSIM} and ID loss \cite{ArcFaceAA}. Lower LPIPS and ID loss value suggests a more realistic synthesized sketch, while higher SCOOT and FSIM values indicate better similarity with artist-drawn sketches. We present the average SCOOT, LPIPS, FSIM and ID loss values across all test samples in Table \ref{table:id}. More details on quantitative evaluations can be found in supplementary material.
\begin{table}[ht]
\centering
\begin{tabular*}{0.75\textwidth}{@{\extracolsep{\fill}}lcccc@{}}
\hline
Method & SCOOT ↑ & LPIPS ↓ & FSIM ↑ & ID ↓                     \\ \hline
HIDA                     & 0.4433                                                           & 0.3214                                                   & 0.3660 & 0.0241                   \\
FSGAN                   & 0.3621                                                        & 0.2890                                                          & 0.3692             & 0.0424         \\
AdaAttN                 & 0.4670                                                           & 0.2600                                                          & 0.3806          & 0.0233            \\
DualStyleGAN                 & 0.4490                                                         & 0.3012                                                        & 0.3631        & 0.0247             \\
Ours         & \textbf{0.5603}                            &                    \textbf{0.2303}                                                & \textbf{0.4283}    & \textbf{0.0206}         \\ \hline
\end{tabular*}
\caption{Quantitative comparison of AdaAttN \cite{AdaAttN}, FSGAN \cite{FS2K}, HIDA \cite{HIDA} and DualStyleGAN \cite{DualStyleGAN} with our method based on SCOOT, LPIPS, FSIM and ID loss. Our method shows considerably better SCOOT and ID loss values indicating more visually appealing and recognizable results. }
\label{table:id}
\end{table}
\subsection{Qualitative Analysis}

\begin{figure*}[t!]
    \centering
    \begin{subfigure}{0.16\textwidth}
        \includegraphics[width=\linewidth]{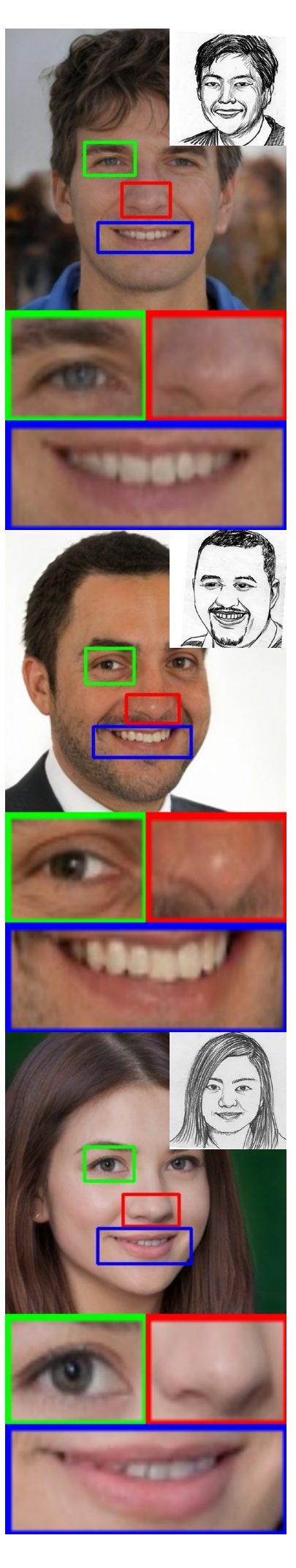}
        
        \label{fig:input_zoom_1}
    \end{subfigure}
    \begin{subfigure}{0.16\textwidth}
        \includegraphics[width=\linewidth]{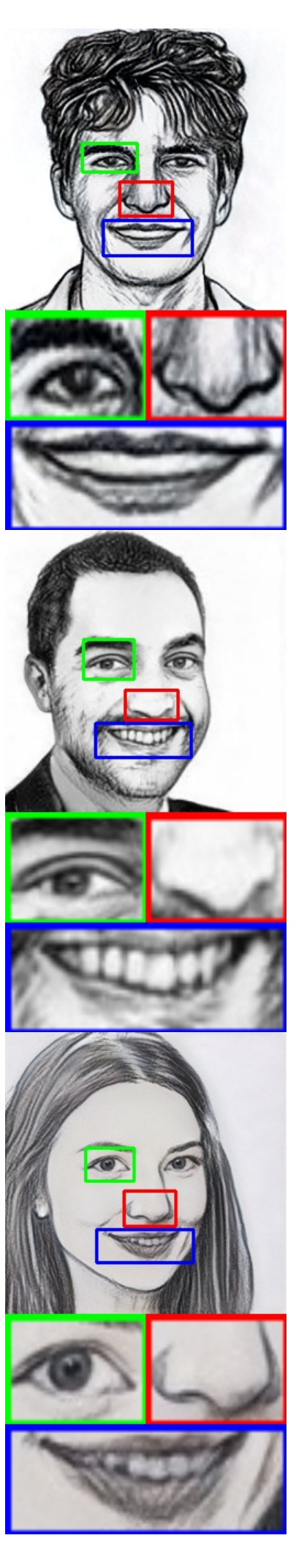}
        
        \label{fig:fs2k_1_zoom_1}
    \end{subfigure}
    \begin{subfigure}{0.16\textwidth}
        \includegraphics[width=\linewidth]{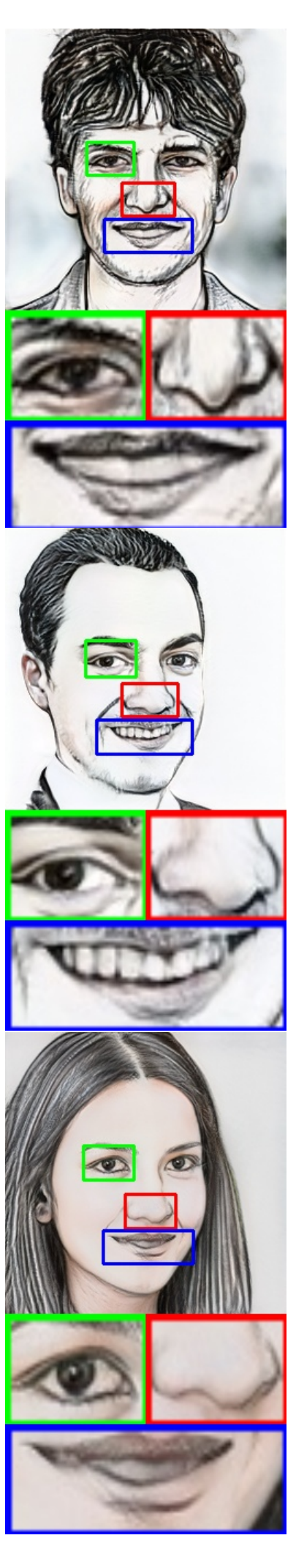}
        
        \label{fig:fs2k_2_zoom_1}
    \end{subfigure}
    \begin{subfigure}{0.16\textwidth}
        \includegraphics[width=\linewidth]{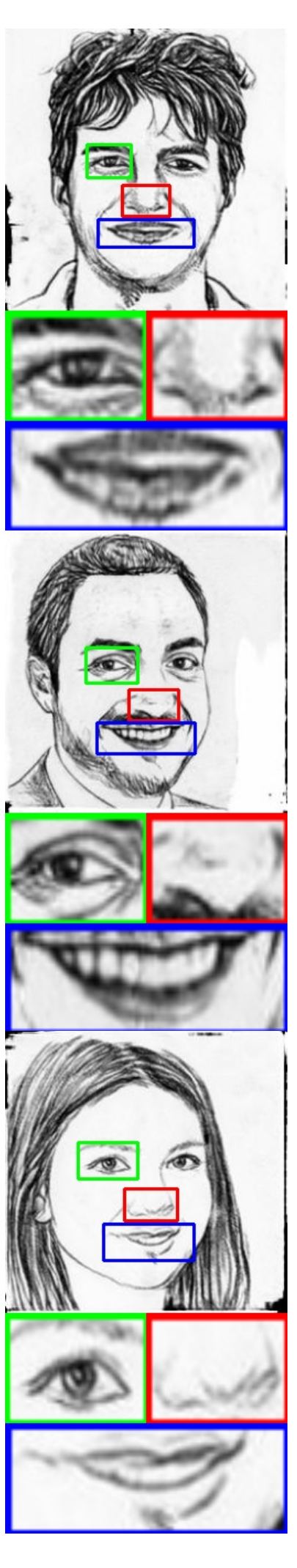}
        
        \label{fig:fs2k_3_zoom_1}
    \end{subfigure}
    \begin{subfigure}{0.16\textwidth}
        \includegraphics[width=\linewidth]{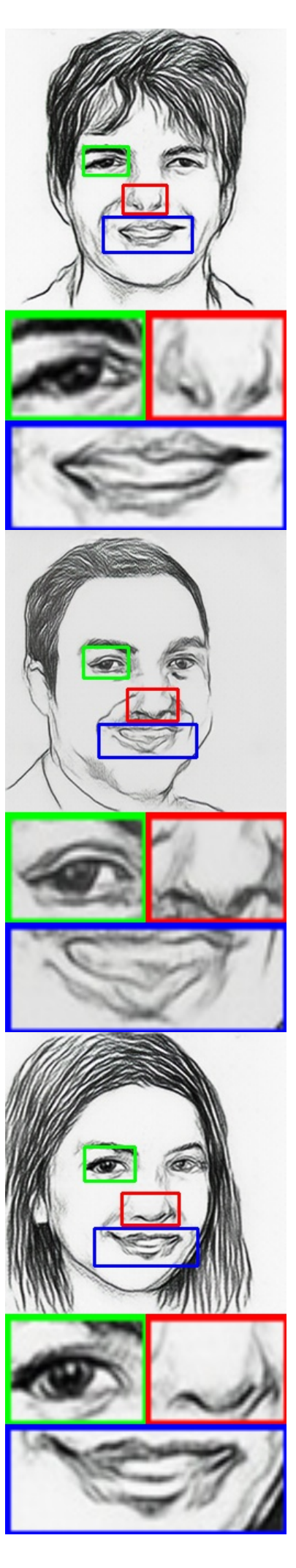}
        
        \label{fig:cuhk_zoom_1}
    \end{subfigure}
    \begin{subfigure}{0.16\textwidth}
        \includegraphics[width=\linewidth]{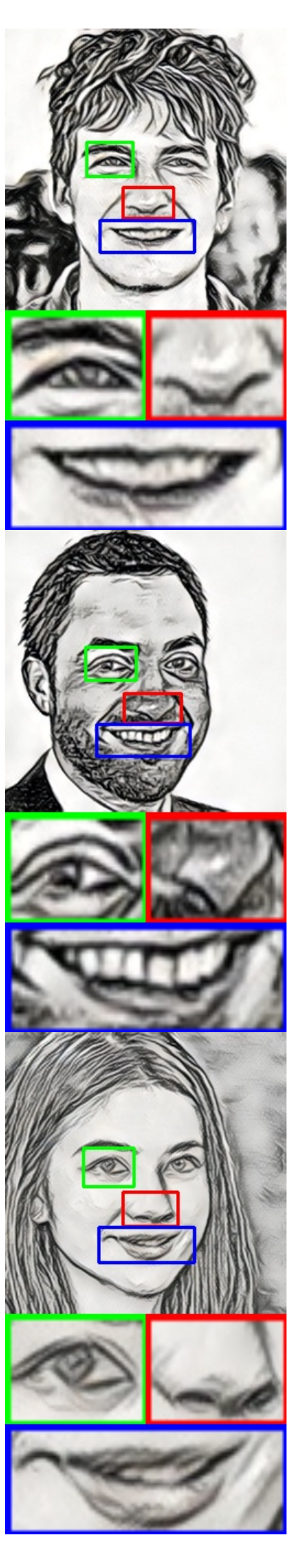}
        
        \label{fig:apd_zoom_1}
    \end{subfigure}
    \caption{Comparison of our method with other state of the art methods on the 3 styles (inset) of FS2K: style 1 (row 1), style 2 (row 2), style 3 (row 3). From left to right : Input identity image, Ours, DualStyleGAN \cite{DualStyleGAN}, HIDA \cite{HIDA}, FSGAN \cite{FS2K}, AdaAttN \cite{AdaAttN}.  }
    \label{fig:comparison}
\end{figure*}

Visually comparing our PS-StyleGAN with leading methods, namely FSGAN \cite{FS2K}, HIDA \cite{HIDA}, DualStyleGAN \cite{DualStyleGAN}, and AdaAttN \cite{AdaAttN}, we observe that our method excels in rendering eyes and lips, showcasing sharper details and enhanced realism, see Fig \ref{fig:comparison}. Our results are visually most similar to DualStyleGAN but their method also learns shape biases in the dataset hence affecting recognizability. DualStyleGAN often changes the gaze direction and shape of lips too. The Attentive Affine transform blocks in PS-StyleGAN contribute to a superior balance between artistic expression and accuracy, resulting in more visually appealing and faithful representations of facial features.

\section{Conclusion}
We introduced \textbf{Portrait Sketching StyleGAN (PS-StyleGAN)}, an approach tailored specifically for the intricate color transformation demands in portrait sketch synthesis. Leveraging the semantic $W+$ latent space of StyleGAN, our method not only generates portrait sketches but also allows meaningful edits, such as pose and expression alterations, while preserving identity. The incorporation of Attentive Affine Transform blocks, fine-tuned through extensive experimentation, allows us to adapt StyleGAN outputs in an inversion-consistent manner by considering both content and style latent features. The model demonstrates efficacy with minimal paired examples (approximately 100) and boasts a short training time, contributing to its practical applicability. However, our method may be susceptible to data bias, and performance could vary across datasets. Additionally, one noteworthy limitation is the current inability to generate realistic accessories in the synthesized sketches. Future work could focus on addressing these limitations to enhance the utility of the proposed PS-StyleGAN further.
\bibliographystyle{splncs04}
\bibliography{main}

\begin{thebibliography}{10}
\providecommand{\url}[1]{\texttt{#1}}
\providecommand{\urlprefix}{URL }
\providecommand{\doi}[1]{https://doi.org/#1}

\bibitem{Abdal_2019_ICCV}
Abdal, R., Qin, Y., Wonka, P.: Image2stylegan: How to embed images into the stylegan latent space? In: Proceedings of the IEEE/CVF International Conference on Computer Vision (ICCV) (October 2019)

\bibitem{DisentangledIG}
Alharbi, Y., Wonka, P.: Disentangled image generation through structured noise injection. 2020 IEEE/CVF Conference on Computer Vision and Pattern Recognition (CVPR) pp. 5133--5141 (2020), \url{https://api.semanticscholar.org/CorpusID:216553495}

\bibitem{style_abstraction}
Berger, I., Shamir, A., Mahler, M., Carter, E.J., Hodgins, J.K.: Style and abstraction in portrait sketching. ACM Transactions on Graphics (TOG)  \textbf{32},  1 -- 12 (2013), \url{https://api.semanticscholar.org/CorpusID:17238299}

\bibitem{SurfaceVE}
Biederman, I., Ju, G.: Surface versus edge-based determinants of visual recognition. Cognitive Psychology  \textbf{20},  38--64 (1988), \url{https://api.semanticscholar.org/CorpusID:14269563}

\bibitem{pix2pixsketch1}
Chan, C., Durand, F., Isola, P.: Learning to generate line drawings that convey geometry and semantics. 2022 IEEE/CVF Conference on Computer Vision and Pattern Recognition (CVPR) pp. 7905--7915 (2022), \url{https://api.semanticscholar.org/CorpusID:247628105}

\bibitem{JoJoGan}
Chong, M.J., Forsyth, D.A.: Jojogan: One shot face stylization. ArXiv  \textbf{abs/2112.11641} (2021), \url{https://api.semanticscholar.org/CorpusID:245385527}

\bibitem{EditingIS}
Collins, E., Bala, R., Price, B., S{\"u}sstrunk, S.: Editing in style: Uncovering the local semantics of gans. 2020 IEEE/CVF Conference on Computer Vision and Pattern Recognition (CVPR) pp. 5770--5779 (2020)

\bibitem{DeCarlo2003SuggestiveCF}
DeCarlo, D., Finkelstein, A., Rusinkiewicz, S., Santella, A.: Suggestive contours for conveying shape. ACM SIGGRAPH 2003 Papers  (2003), \url{https://api.semanticscholar.org/CorpusID:1485904}

\bibitem{ArcFaceAA}
Deng, J., Guo, J., Zafeiriou, S.: Arcface: Additive angular margin loss for deep face recognition. 2019 IEEE/CVF Conference on Computer Vision and Pattern Recognition (CVPR) pp. 4685--4694 (2018)

\bibitem{dhariwal2021diffusion}
Dhariwal, P., Nichol, A.Q.: Diffusion models beat {GAN}s on image synthesis. In: Beygelzimer, A., Dauphin, Y., Liang, P., Vaughan, J.W. (eds.) Advances in Neural Information Processing Systems (2021), \url{https://openreview.net/forum?id=AAWuCvzaVt}

\bibitem{FS2K}
Fan, D.P., Huang, Z., Zheng, P., Liu, H., Qin, X., Gool, L.V.: Facial-sketch synthesis: A new challenge. Machine Intelligence Research  \textbf{19},  257 -- 287 (2021), \url{https://api.semanticscholar.org/CorpusID:248987735}

\bibitem{Scoot}
Fan, D.P., Zhang, S., Wu, Y.H., Liu, Y., Cheng, M.M., Ren, B., Rosin, P.L., Ji, R.: Scoot: A perceptual metric for facial sketches. In: The IEEE International Conference on Computer Vision (ICCV) (2019)

\bibitem{Gal2022TI}
Gal, R., Alaluf, Y., Atzmon, Y., Patashnik, O., Bermano, A.H., Chechik, G., Cohen-Or, D.: An image is worth one word: Personalizing text-to-image generation using textual inversion. ArXiv  \textbf{abs/2208.01618} (2022), \url{https://api.semanticscholar.org/CorpusID:251253049}

\bibitem{HIDA}
Gao, F., Zhu, Y., Jiang, C., Wang, N.: Human-inspired facial sketch synthesis with dynamic adaptation. In: Proceedings of the International Conference on Computer Vision (ICCV) (2023)

\bibitem{Gatys2016ImageST}
Gatys, L.A., Ecker, A.S., Bethge, M.: Image style transfer using convolutional neural networks. 2016 IEEE Conference on Computer Vision and Pattern Recognition (CVPR) pp. 2414--2423 (2016), \url{https://api.semanticscholar.org/CorpusID:206593710}

\bibitem{Goodfellow2014GenerativeAN}
Goodfellow, I.J., Pouget-Abadie, J., Mirza, M., Xu, B., Warde-Farley, D., Ozair, S., Courville, A.C., Bengio, Y.: Generative adversarial nets. In: Neural Information Processing Systems (2014), \url{https://api.semanticscholar.org/CorpusID:261560300}

\bibitem{Unsupervised_editing}
H{\"a}rk{\"o}nen, E., Hertzmann, A., Lehtinen, J., Paris, S.: Ganspace: Discovering interpretable gan controls. ArXiv  \textbf{abs/2004.02546} (2020), \url{https://api.semanticscholar.org/CorpusID:214802845}

\bibitem{ResNet}
He, K., Zhang, X., Ren, S., Sun, J.: Deep residual learning for image recognition. 2016 IEEE Conference on Computer Vision and Pattern Recognition (CVPR) pp. 770--778 (2015), \url{https://api.semanticscholar.org/CorpusID:206594692}

\bibitem{HertzmannWhyDL}
Hertzmann, A.: Why do line drawings work? a realism hypothesis. Perception  \textbf{49},  439 -- 451 (2020), \url{https://api.semanticscholar.org/CorpusID:211132554}

\bibitem{HertzmannROLE}
Hertzmann, A.: The role of edges in line drawing perception. Perception  \textbf{50},  266 -- 275 (2021), \url{https://api.semanticscholar.org/CorpusID:231698870}

\bibitem{Ho2020DenoisingDP}
Ho, J., Jain, A., Abbeel, P.: Denoising diffusion probabilistic models. ArXiv  \textbf{abs/2006.11239} (2020), \url{https://api.semanticscholar.org/CorpusID:219955663}

\bibitem{StyleTF}
Hu, X., Huang, Q., Shi, Z., Li, S., Gao, C., Sun, L., Li, Q.: Style transformer for image inversion and editing. 2022 IEEE/CVF Conference on Computer Vision and Pattern Recognition (CVPR) pp. 11327--11336 (2022), \url{https://api.semanticscholar.org/CorpusID:247450902}

\bibitem{AdaIN}
Huang, X., Belongie, S.J.: Arbitrary style transfer in real-time with adaptive instance normalization. 2017 IEEE International Conference on Computer Vision (ICCV) pp. 1510--1519 (2017), \url{https://api.semanticscholar.org/CorpusID:6576859}

\bibitem{pix2pix2017}
Isola, P., Zhu, J.Y., Zhou, T., Efros, A.A.: Image-to-image translation with conditional adversarial networks. CVPR  (2017)

\bibitem{jain2024clip4sketchenhancingsketchmugshot}
Jain, K.K., Grosz, S., Namboodiri, A.M., Jain, A.K.: Clip4sketch: Enhancing sketch to mugshot matching through dataset augmentation using diffusion models (2024), \url{https://arxiv.org/abs/2408.01233}

\bibitem{perceptual}
Johnson, J., Alahi, A., Fei-Fei, L.: Perceptual losses for real-time style transfer and super-resolution. In: Computer Vision--ECCV 2016: 14th European Conference, Amsterdam, The Netherlands, October 11-14, 2016, Proceedings, Part II 14. pp. 694--711. Springer (2016)

\bibitem{Karras2018ASG}
Karras, T., Laine, S., Aila, T.: A style-based generator architecture for generative adversarial networks. 2019 IEEE/CVF Conference on Computer Vision and Pattern Recognition (CVPR) pp. 4396--4405 (2018), \url{https://api.semanticscholar.org/CorpusID:54482423}

\bibitem{Karras_2019_CVPR}
Karras, T., Laine, S., Aila, T.: A style-based generator architecture for generative adversarial networks. In: Proceedings of the IEEE/CVF Conference on Computer Vision and Pattern Recognition (CVPR) (June 2019)

\bibitem{UGATIT}
Kim, J., Kim, M., Kang, H., Lee, K.H.: U-gat-it: Unsupervised generative attentional networks with adaptive layer-instance normalization for image-to-image translation. In: International Conference on Learning Representations (2019)

\bibitem{TransformingTL}
Li, H., Liu, J., Bai, Y., Wang, H., Mueller, K.: Transforming the latent space of stylegan for real face editing. The Visual Computer pp. 1--16 (2021), \url{https://api.semanticscholar.org/CorpusID:235254637}

\bibitem{li2023photomaker}
Li, Z., Cao, M., Wang, X., Qi, Z., Cheng, M.M., Shan, Y.: Photomaker: Customizing realistic human photos via stacked id embedding. In: IEEE Conference on Computer Vision and Pattern Recognition (CVPR) (2024)

\bibitem{Hertsmann_stroke_3d}
Liu, D., Fisher, M., Hertzmann, A., Kalogerakis, E.: Neural strokes: Stylized line drawing of 3d shapes. 2021 IEEE/CVF International Conference on Computer Vision (ICCV) pp. 14184--14193 (2021), \url{https://api.semanticscholar.org/CorpusID:238531682}

\bibitem{UNIT}
Liu, M.Y., Breuel, T.M., Kautz, J.: Unsupervised image-to-image translation networks. In: Neural Information Processing Systems (2017), \url{https://api.semanticscholar.org/CorpusID:3783306}

\bibitem{AdaAttN}
Liu, S., Lin, T., He, D., Li, F., Wang, M., Li, X., Sun, Z., Li, Q., Ding, E.: Adaattn: Revisit attention mechanism in arbitrary neural style transfer. 2021 IEEE/CVF International Conference on Computer Vision (ICCV) pp. 6629--6638 (2021), \url{https://api.semanticscholar.org/CorpusID:236956663}

\bibitem{contextual}
Mechrez, R., Talmi, I., Zelnik-Manor, L.: The contextual loss for image transformation with non-aligned data. In: Proceedings of the European conference on computer vision (ECCV). pp. 768--783 (2018)

\bibitem{cGAN}
Mirza, M., Osindero, S.: Conditional generative adversarial nets. ArXiv  \textbf{abs/1411.1784} (2014)

\bibitem{OhtakeRidge}
Ohtake, Y., Belyaev, A.G., Seidel, H.P.: Ridge-valley lines on meshes via implicit surface fitting. ACM SIGGRAPH 2004 Papers  (2004), \url{https://api.semanticscholar.org/CorpusID:8500135}

\bibitem{SANet}
Park, D.Y., Lee, K.H.: Arbitrary style transfer with style-attentional networks. 2019 IEEE/CVF Conference on Computer Vision and Pattern Recognition (CVPR) pp. 5873--5881 (2018), \url{https://api.semanticscholar.org/CorpusID:54447797}

\bibitem{clip}
Radford, A., Kim, J.W., Hallacy, C., Ramesh, A., Goh, G., Agarwal, S., Sastry, G., Askell, A., Mishkin, P., Clark, J., Krueger, G., Sutskever, I.: Learning transferable visual models from natural language supervision. In: International Conference on Machine Learning (2021), \url{https://api.semanticscholar.org/CorpusID:231591445}

\bibitem{Ramesh2022HierarchicalTI}
Ramesh, A., Dhariwal, P., Nichol, A., Chu, C., Chen, M.: Hierarchical text-conditional image generation with clip latents. ArXiv  \textbf{abs/2204.06125} (2022), \url{https://api.semanticscholar.org/CorpusID:248097655}

\bibitem{Richardson2020EncodingIS}
Richardson, E., Alaluf, Y., Patashnik, O., Nitzan, Y., Azar, Y., Shapiro, S., Cohen-Or, D.: Encoding in style: a stylegan encoder for image-to-image translation. 2021 IEEE/CVF Conference on Computer Vision and Pattern Recognition (CVPR) pp. 2287--2296 (2020)

\bibitem{Rombach2021HighResolutionIS}
Rombach, R., Blattmann, A., Lorenz, D., Esser, P., Ommer, B.: High-resolution image synthesis with latent diffusion models. 2022 IEEE/CVF Conference on Computer Vision and Pattern Recognition (CVPR) pp. 10674--10685 (2021), \url{https://api.semanticscholar.org/CorpusID:245335280}

\bibitem{Ruiz2022DreamBoothFT}
Ruiz, N., Li, Y., Jampani, V., Pritch, Y., Rubinstein, M., Aberman, K.: Dreambooth: Fine tuning text-to-image diffusion models for subject-driven generation. 2023 IEEE/CVF Conference on Computer Vision and Pattern Recognition (CVPR) pp. 22500--22510 (2022), \url{https://api.semanticscholar.org/CorpusID:251800180}

\bibitem{NPRLineDF}
Rusinkiewicz, S., DeCarlo, D., Finkelstein, A.: Line drawings from 3d models. In: International Conference on Computer Graphics and Interactive Techniques (2005), \url{https://api.semanticscholar.org/CorpusID:10994464}

\bibitem{Saharia2022PhotorealisticTD}
Saharia, C., Chan, W., Saxena, S., Li, L., Whang, J., Denton, E.L., Ghasemipour, S.K.S., Ayan, B.K., Mahdavi, S.S., Lopes, R.G., Salimans, T., Ho, J., Fleet, D.J., Norouzi, M.: Photorealistic text-to-image diffusion models with deep language understanding. ArXiv  \textbf{abs/2205.11487} (2022), \url{https://api.semanticscholar.org/CorpusID:248986576}

\bibitem{Linesasedges}
Sayim, B., Cavanagh, P.: What line drawings reveal about the visual brain. Frontiers in Human Neuroscience  \textbf{5} (2011), \url{https://api.semanticscholar.org/CorpusID:263708652}

\bibitem{VeryDC}
Simonyan, K., Zisserman, A.: Very deep convolutional networks for large-scale image recognition. CoRR  \textbf{abs/1409.1556} (2014), \url{https://api.semanticscholar.org/CorpusID:14124313}

\bibitem{AgileGAN}
Song, G., Luo, L., Liu, J., Ma, W.C., Lai, C.P., Zheng, C., Cham, T.J.: Agilegan: stylizing portraits by inversion-consistent transfer learning. ACM Trans. Graph.  \textbf{40},  117:1--117:13 (2021), \url{https://api.semanticscholar.org/CorpusID:236006017}

\bibitem{E4E}
Tov, O., Alaluf, Y., Nitzan, Y., Patashnik, O., Cohen-Or, D.: Designing an encoder for stylegan image manipulation. ACM Transactions on Graphics (TOG)  \textbf{40},  1 -- 14 (2021), \url{https://api.semanticscholar.org/CorpusID:231802331}

\bibitem{MechanismsOF}
Tsao, D.Y., Livingstone, M.S.: Mechanisms of face perception. Annual review of neuroscience  \textbf{31},  411--37 (2008), \url{https://api.semanticscholar.org/CorpusID:14760952}

\bibitem{Ulyanov2016texture}
Ulyanov, D., Lebedev, V., Vedaldi, A., Lempitsky, V.: Texture networks: Feed-forward synthesis of textures and stylized images. In: 33rd International Conference on Machine Learning, ICML 2016. pp. 2027--2041 (2016)

\bibitem{Attention}
Vaswani, A., Shazeer, N.M., Parmar, N., Uszkoreit, J., Jones, L., Gomez, A.N., Kaiser, L., Polosukhin, I.: Attention is all you need. In: Neural Information Processing Systems (2017), \url{https://api.semanticscholar.org/CorpusID:13756489}

\bibitem{cuhk}
Wang, X., Tang, X.: Face photo-sketch synthesis and recognition. IEEE Transactions on Pattern Analysis and Machine Intelligence  \textbf{31}(11),  1955--1967 (2009). \doi{10.1109/TPAMI.2008.222}

\bibitem{TransEditor}
Xu, Y., Yin, Y., Jiang, L., Wu, Q., Zheng, C., Loy, C.C., Dai, B., Wu, W.: Transeditor: Transformer-based dual-space gan for highly controllable facial editing. 2022 IEEE/CVF Conference on Computer Vision and Pattern Recognition (CVPR) pp. 7673--7682 (2022), \url{https://api.semanticscholar.org/CorpusID:247839345}

\bibitem{DualStyleGAN}
Yang, S., Jiang, L., Liu, Z., Loy, C.C.: Pastiche master: Exemplar-based high-resolution portrait style transfer. 2022 IEEE/CVF Conference on Computer Vision and Pattern Recognition (CVPR) pp. 7683--7692 (2022), \url{https://api.semanticscholar.org/CorpusID:247627720}

\bibitem{Ye2023IPAdapterTC}
Ye, H., Zhang, J., Liu, S., Han, X., Yang, W.: Ip-adapter: Text compatible image prompt adapter for text-to-image diffusion models. ArXiv  \textbf{abs/2308.06721} (2023), \url{https://api.semanticscholar.org/CorpusID:260886966}

\bibitem{APDrawingGAN2019}
Yi, R., Liu, Y.J., Lai, Y.K., Rosin, P.L.: Apdrawinggan: Generating artistic portrait drawings from face photos with hierarchical gans. 2019 IEEE/CVF Conference on Computer Vision and Pattern Recognition (CVPR) pp. 10735--10744 (2019), \url{https://api.semanticscholar.org/CorpusID:194358484}

\bibitem{FSIM}
Zhang, L., Zhang, L., Mou, X., Zhang, D.: Fsim: A feature similarity index for image quality assessment. IEEE Transactions on Image Processing  \textbf{20},  2378--2386 (2011), \url{https://api.semanticscholar.org/CorpusID:10649298}

\bibitem{lpips}
Zhang, R., Isola, P., Efros, A.A., Shechtman, E., Wang, O.: The unreasonable effectiveness of deep features as a perceptual metric. 2018 IEEE/CVF Conference on Computer Vision and Pattern Recognition pp. 586--595 (2018), \url{https://api.semanticscholar.org/CorpusID:4766599}

\bibitem{cyclegan}
Zhu, J.Y., Park, T., Isola, P., Efros, A.A.: Unpaired image-to-image translation using cycle-consistent adversarial networks. 2017 IEEE International Conference on Computer Vision (ICCV) pp. 2242--2251 (2017)

\end{thebibliography}
%




\end{document}